\documentclass[runningheads]{llncs}
\usepackage{lmodern}
\usepackage{graphicx,booktabs,array,amsmath,amssymb,microtype,tabularx}
\usepackage{tikz}
\usetikzlibrary{positioning,arrows.meta}
\usepackage{enumitem}
\usepackage{listings}
\usepackage{float,needspace}
\usepackage[hidelinks]{hyperref}
\setlist[itemize]{leftmargin=*,nosep}
\setlist[enumerate]{leftmargin=*,nosep}
\newcommand{\preprintstatus}{\begin{center}\small\textit{Preprint. 22 October 2026.}\end{center}}
\hypersetup{pdftitle={Class-Grounded Batik Captioning under Scarce Text Annotation: Label Retention and Source-Specific Error Propagation},pdfauthor={Novanto Yudistira; Muhammad Dzaka Aufa Fadhillah; Irawati Nurmala Sari}}
\begin{document}

\title{Class-Grounded Batik Captioning under Scarce Text Annotation: Label Retention and Source-Specific Error Propagation}
\titlerunning{Label Retention and Error Propagation in Batik Captioning}
\author{Novanto Yudistira \and Muhammad Dzaka Aufa Fadhillah \and Irawati Nurmala Sari}
\authorrunning{N. Yudistira et al.}
\institute{Universitas Brawijaya, Malang, Indonesia}
\maketitle
\preprintstatus

\begin{abstract}
Class labels offer compact supervision for describing cultural-heritage images when expert text annotations are scarce. We investigate a batik-captioning cascade that integrates a LoRA-adapted DINOv2 classifier with Qwen2.5-VL-7B, adapted using LoRA and loaded in 4-bit NF4. A $3\times2$ evaluation combines three captioner conditions with two inference prompts to examine adaptation and predicted-class injection. Across 150 caption-evaluation images, all three injected conditions achieve a ground-truth class-label mention rate of 0.9067. The classifier's separate test yields 96.10\% pooled accuracy, with 98.33\% on 300 public-source images and 75.76\% on 33 local images. Supported-class macro-F1 and balanced accuracy on the local subset are 0.8493 and 0.8302, respectively. The classifier and captioner use separate partitions, and their training--evaluation overlap remains unquantified. Differences in label semantics and training support also affect interpretation of the source-specific accuracy gap. Reference-based caption metrics and image--text compatibility scores produce different model rankings. Qualitative examples show both retention of correctly predicted identities and propagation of classification errors into fluent descriptions. These findings characterize how class conditioning preserves label information and transfers upstream errors into batik captions under scarce text annotation.
\end{abstract}

\section{Introduction}
Cultural-heritage digitization frequently produces image collections with category labels and limited descriptive text. Batik archives exemplify this setting: a catalogue may identify a regional class or motif while providing sparse information about visible colors, ornaments, and spatial arrangement. Generating useful descriptions therefore requires a model to connect visual appearance with collection-specific terminology.

\begin{figure}[H]\centering
\begin{minipage}[t]{0.315\textwidth}\centering\small
\textbf{(a) Tasik Madu}\par\smallskip
\parbox[c][3.5cm][c]{\linewidth}{\centering\includegraphics[width=\linewidth,height=3.5cm,keepaspectratio]{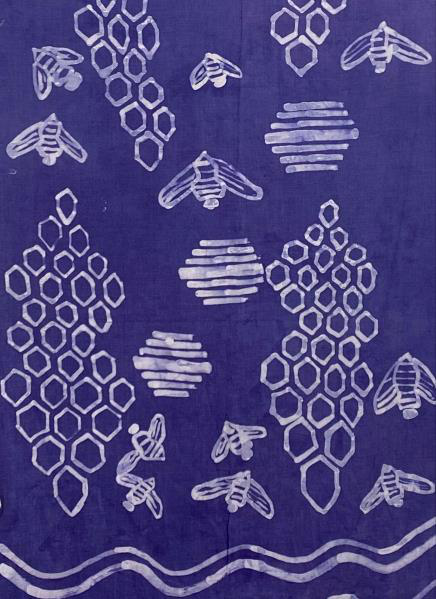}}\par\smallskip
\textbf{Annotated class: Malang}\par\medskip
\raggedright\footnotesize
\textbf{Generated caption excerpt:}\par
``the batik fabric features the \textbf{malang} motif, characterized by white motifs against a deep blue background. the dominant motif elements include stylized bees and honeycomb structures.''
\end{minipage}\hfill
\begin{minipage}[t]{0.315\textwidth}\centering\small
\textbf{(b) Bunga Nirwana 2}\par\smallskip
\parbox[c][3.5cm][c]{\linewidth}{\centering\includegraphics[width=\linewidth,height=3.5cm,keepaspectratio]{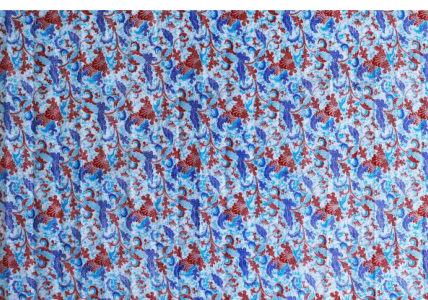}}\par\smallskip
\textbf{Annotated class: Tulungagung}\par\medskip
\raggedright\footnotesize
\textbf{Generated caption excerpt:}\par
``the batik fabric features the \textbf{trenggalek} motif, characterized by red, blue, and white motifs. the dominant motif elements appear to be stylized leaves and flowers.''
\end{minipage}\hfill
\begin{minipage}[t]{0.315\textwidth}\centering\small
\textbf{(c) Melati Deret}\par\smallskip
\parbox[c][3.5cm][c]{\linewidth}{\centering\includegraphics[width=\linewidth,height=3.5cm,keepaspectratio]{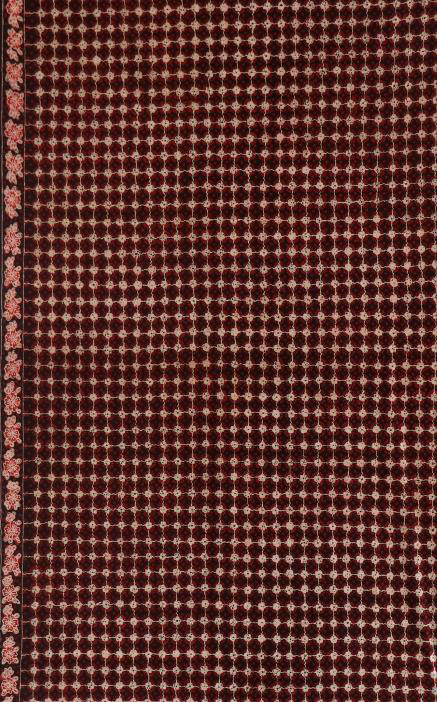}}\par\smallskip
\textbf{Annotated class: Lamongan}\par\medskip
\raggedright\footnotesize
\textbf{Generated caption excerpt:}\par
``the batik fabric features the \textbf{trenggalek} motif, characterized by deep red motifs. the dominant motif elements appear to be small, intricate designs that resemble traditional indonesian patterns.''
\end{minipage}
\caption{Batik images and generated caption excerpts from Dataset A. Bold class names indicate the model's cultural attribution. The Malang identity in (a) agrees with the annotation; the Trenggalek identities in (b) and (c) conflict with the annotated Tulungagung and Lamongan classes. All outputs use the captioner adapted with class labels and a predicted-class inference prompt (CIFT-CI, defined in Section~\ref{sec:adaptation}).}
\label{fig:examples}
\end{figure}

\Needspace{32\baselineskip}
Figure~\ref{fig:examples} introduces three batik images with their annotated regional classes and caption excerpts generated using predicted-class conditioning. The examples connect the captioning task to visible fabric patterns and illustrate how a classifier prediction can determine the cultural identity expressed in the text.

A vision--language model (VLM) can generate fluent descriptions, whereas reliable cultural attribution depends on recognizing the relevant class. These requirements motivate a two-stage framework in which a visual classifier predicts a cultural class and a captioner uses that prediction as contextual information. Previous studies have examined batik recognition and synthesis \cite{gultom2018batik,octadion2025batik,wahid2025batik}, culturally informed captioning \cite{yun2024cic}, and classifier-guided radiology captioning \cite{chowdhury2025modality}.

Class conditioning also links the errors of the two stages. A correct prediction can introduce a collection-specific identity into the caption, while an incorrect prediction can produce a fluent description with an erroneous identity. Evaluation consequently needs to distinguish label retention, agreement with reference text, visual compatibility, and factual correctness. Repeated reference templates introduce a further challenge because lexical similarity can reward recurring expressions across visually different images.

We investigate these effects through a DINOv2--Qwen2.5-VL cascade. The evaluation crosses three captioner conditions with two inference prompts and examines class-label mention, caption metrics, classifier performance, and qualitative error patterns. The study contributes:
\begin{itemize}
\item a six-condition evaluation of captioner adaptation and inference-time class injection under scarce text annotation;
\item an analysis of source-specific and class-specific classifier performance with explicit metric-averaging conventions; and
\item an examination of label retention, caption-metric disagreement, downstream identity errors, and resource use.
\end{itemize}
The evaluation focuses on the behavior of class-conditioned caption generation under independently constructed classifier and captioner partitions. Section~\ref{sec:evaluation} defines this evaluation scope, and Section~\ref{sec:discussion} discusses its implications for complete-cascade generalization.

\section{Related Work}
Batik recognition research has explored transfer learning \cite{gultom2018batik}, while diffusion and GAN-based approaches address motif synthesis \cite{octadion2025batik,wahid2025batik}. Indonesian image captioning has also been studied using ViT--IndoBERT on general image--text data \cite{apandi2025indocap}. Our study examines the connection between cultural-class recognition and generated descriptions of batik images.

Culturally aware captioning incorporates external contextual information \cite{yun2024cic}. Classifier-guided captioning provides a related interface by passing compact domain information to a language-generation stage \cite{chowdhury2025modality}. We use this interface to analyze the retention of predicted identities and the propagation of classification errors.

DINOv2 provides transferable visual representations \cite{oquab2024dinov2}, and Qwen2-VL and Qwen2.5-VL support multimodal instruction following \cite{wang2024qwen2vl,bai2025qwen25vl}. LoRA reduces the number of parameters updated during adaptation \cite{hu2022lora}, while quantized adaptation supports lower-memory model loading \cite{dettmers2023qlora}. These components enable the classifier and captioner to be adapted separately within the cascade.

Reference-based metrics quantify lexical or semantic agreement with target captions \cite{papineni2002bleu,lin2004rouge,banerjee2005meteor,anderson2016spice,zhang2020bertscore}. CLIPScore evaluates image--text compatibility \cite{hessel2021clipscore}. Research on hallucination further motivates direct examination of the visual assertions made in generated captions \cite{rohrbach2018hallucination,li2023pope}. Together, these perspectives support evaluating cultural identity and visible attributes as distinct aspects of caption content.

\section{Data and Task}
\subsection{Image collections and cultural labels}
Dataset A contains 346 locally curated East-Java images assigned to four regional classes: Lamongan, Malang, Trenggalek, and Tulungagung. Human-curated metadata describes the class, colors, visually grounded theme, and pattern shape. Reference captions are constructed from these fields using a recurring template.\footnote{The evaluation examples use the following reference pattern: \lq\lq The batik fabric features the [class] motif, characterized by [colors] motifs. The dominant motifs include [theme]. The motifs are arranged in a [shape] pattern across the fabric, creating a harmonious and balanced composition.\rq\rq} The reference text is therefore template-based, with descriptive content limited to the selected metadata fields. Historical interpretation, symbolism, fabric type, dyes, and production techniques are excluded from the caption target.

Dataset B comprises 3,000 images from 20 classes, with 150 images per class. The data documentation identifies Kaggle as the host platform; the repository identifier, creator, release, and license remain unspecified. This incomplete provenance limits reconstruction of the collection. Appendix~\ref{app:labels} lists the 20 class names.

The combined inventory contains 24 cultural classes. Dataset A uses regional labels, while Dataset B primarily uses motif names together with several regional categories. Both collections contribute to training. Comparisons between sources consequently reflect differences in label semantics, sample support, and potentially acquisition conditions.

\Needspace{16\baselineskip}
Table~\ref{tab:data} summarizes the collection sizes, label inventories, supervision, and component-specific data allocations.

\begin{table}[H]
\caption{Data allocation and supervision for the classifier and captioner. The two components use separately constructed partitions.}
\label{tab:data}\centering\small
\begin{tabular}{lrrp{4.7cm}}
\toprule
Source & Images & Labels & Allocation / supervision \\
\midrule
A & 346 & 4 & Captioner: 163 train, 33 validation, 150 test; metadata-derived references. Classifier test: 33. \\
B & 3,000 & 20 & Captioner: 2,700 train, 300 validation; pseudo-captions. Classifier test: 300. \\
\bottomrule
\end{tabular}
\end{table}

\Needspace{23\baselineskip}
\subsection{Class-conditioned caption generation}
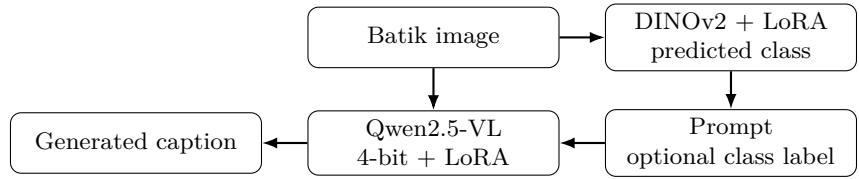
\begin{figure}[H]\centering
\begin{tikzpicture}[node distance=5mm and 6mm,
box/.style={draw,rounded corners,align=center,minimum height=8mm,text width=31mm,font=\small},
arr/.style={-{Latex[length=2mm]},thick}]
\node[box] (image) {Batik image};
\node[box,right=of image] (classifier) {DINOv2 + LoRA\\predicted class};
\node[box,below=of classifier] (prompt) {Prompt\\optional class label};
\node[box,left=of prompt] (vlm) {Qwen2.5-VL\\4-bit + LoRA};
\node[box,left=of vlm] (caption) {Generated caption};
\draw[arr] (image)--(classifier);
\draw[arr] (classifier)--(prompt);
\draw[arr] (prompt)--(vlm);
\draw[arr] (image)--(vlm);
\draw[arr] (vlm)--(caption);
\end{tikzpicture}
\caption{Batik-captioning cascade with optional class injection. The classifier prediction provides contextual information to the captioner and forms a pathway for downstream identity errors.}
\label{fig:pipeline}
\end{figure}

Figure~\ref{fig:pipeline} illustrates the classifier--captioner cascade and its optional class input. For image $x$, the classifier predicts $\hat y=\arg\max_y p_\theta(y\mid x)$. The captioner generates $c=g_\phi(x,p)$ using either a generic prompt $p_0$ in the no-class-injection (NCI) condition or a class-conditioned prompt $p(\hat y)$ in the class-injection (CI) condition. Paired prompt conditions use the same image and captioner. The resulting comparison characterizes the effect of predicted-class conditioning on generated text.

\section{Methodology}
\subsection{Pseudo-caption generation}
For Dataset B, zero-shot pseudo-captions are generated with Qwen2.5-VL-7B-Instruct conditioned on each image and its class label before domain adaptation. The generation prompt requests dominant colors, visible elements, spatial arrangement, and visible \emph{isen-isen}, while excluding unsupported historical or symbolic interpretations. It also requires the class label in the opening sentence (Appendix~\ref{app:prompt}), directly encouraging identity mention.

Manual review removes or regenerates captions with incorrect colors, class-inconsistent descriptions, or unsupported cultural assertions. The resulting corpus contains 3,000 image--caption pairs, comprising 2,700 training pairs and 300 validation pairs.

Pseudo-caption generation disables sampling and uses a maximum of 256 new tokens and a repetition penalty of 1.1. Inference-time caption generation enables sampling with a temperature of 0.3 and a maximum of 256 new tokens.

\subsection{Classifier and captioner adaptation}\label{sec:adaptation}
The DINOv2 backbone is frozen except for LoRA modules and a trained 24-way linear classification head. LoRA targets the query, key, value, and dense projections, using rank 32, scaling 64, and dropout 0.1. Training uses AdamW \cite{loshchilov2019adamw}, a learning rate of $3\times10^{-4}$, weight decay of $10^{-4}$, 10 epochs, cosine scheduling, 10\% warmup, batch size 16, and gradient accumulation over two batches. The objective is a weighted focal-style loss with $\gamma=2$ \cite{lin2017focal}.

Qwen2.5-VL-7B-Instruct is loaded with NF4 and double quantization. LoRA targets the $\{q,k,v,o,gate,up,down\}$ projections, with rank 16, scaling 32, and dropout 0.05. Adaptation uses AdamW, a learning rate of $10^{-4}$, 10 epochs, cosine scheduling, 5\% warmup, batch size 8, and gradient accumulation over four batches. Instruction and image tokens are masked from the answer-token cross-entropy objective. Dynamic-resolution processing uses a 262,144-pixel ceiling.

The three captioner conditions are \textbf{Baseline}, the pretrained model; \textbf{PFT}, adapted using training prompts that omit class labels; and \textbf{CIFT}, adapted using training prompts that include class labels. Crossing these conditions with NCI and CI produces the six-condition evaluation.

\section{Experimental Setup and Evaluation Protocol}\label{sec:evaluation}
\subsection{Data partitions and evaluation scope}
The classifier uses an approximately 80/10/10 stratified split of the combined collection, with 333 test images. The captioner uses the allocations in Table~\ref{tab:data}, including 150 Dataset-A images for caption evaluation. These partitions were constructed independently, creating a risk that the classifier encountered some caption-evaluation images during training. Image-level partition assignments and duplicate-group identifiers are unavailable, leaving the extent of this overlap unquantified.

The 150-image caption evaluation characterizes generation under the implemented cascade protocol. Its interpretation as complete-cascade generalization requires a shared partition that excludes evaluation images from both components. This also requires exact- and near-duplicate grouping before assignment, followed by training and model selection using the corresponding training and validation partitions.

\subsection{Evaluation metrics}
Classifier accuracy measures the fraction of correct class predictions. Supported-class macro-F1 averages per-class F1 over labels with positive ground-truth support in each evaluated source. Predictions into labels outside that supported set remain classification errors. Balanced accuracy averages recall over the supported labels. Class supports and the averaging set are reported alongside the aggregate metrics.

BLEU-1/4, ROUGE-L, METEOR, SPICE, and BERTScore quantify agreement with the reference captions. The recurring reference template makes these measures sensitive to shared syntax and repeated nouns. Metric-library versions, tokenization, BLEU smoothing and aggregation, the BERTScore backbone, and normalization details remain unspecified, limiting independent reproduction of the descriptive scores.

\emph{Motif Mention Rate} (\textbf{MMR}) quantifies ground-truth class-label mention:
\begin{equation}
\mathrm{MMR}=\frac1n\sum_{i=1}^{n}\mathbf1\{\text{class string }y_i\text{ occurs in caption }c_i\}.
\end{equation}
Each image contributes a binary outcome. Under class injection, MMR reflects both upstream label correctness and the captioner's retention of the predicted label. Its scope is class-string inclusion; factual evaluation of the remaining caption content requires attribute-level judgments. Reproduction of MMR also requires explicit string normalization and class-alias rules.

CLIPScore is used as an image--text compatibility diagnostic. Its encoder checkpoint and score scaling remain unspecified. Rare local class names can be weakly represented in the embedding space, so interpretation emphasizes score disagreement alongside direct inspection of caption content.

\subsection{Computing environment and variability}
The computing environment comprises an NVIDIA L40, an AMD EPYC 7413, PyTorch 2.5.1, Transformers 4.52.1, PEFT 0.10.0, and BitsAndBytes 0.49.2. Latency is measured on 30 images and summarized using component means and standard deviations across images. These standard deviations describe image-level runtime variability. Complete per-image caption outputs and repeated-training summaries are unavailable; the analysis consequently uses descriptive comparisons, with paired uncertainty and training-run variability remaining open evaluation requirements.

\Needspace{20\baselineskip}
\section{Results}\label{sec:results}
\subsection{Classifier performance across sources}
Table~\ref{tab:classifier} reports pooled and source-specific classifier accuracy. The classifier reaches pooled accuracy of 96.10\%. Dataset B contributes 300 of the 333 test images and attains 98.33\% accuracy, while Dataset A attains 75.76\% on 33 images.

\begin{table}[H]\centering\small
\caption{Classifier accuracy by source. Both image sources are represented during training, and the source subsets use different cultural-label inventories.}
\label{tab:classifier}
\begin{tabular}{lrrr}
\toprule
Subset & Test images & Correct & Accuracy \\
\midrule
Pooled & 333 & 320 & 0.9610 \\
B & 300 & 295 & 0.9833 \\
A & 33 & 25 & 0.7576 \\
\bottomrule
\end{tabular}
\end{table}

The difference between the displayed source-specific percentages is 22.57 percentage points. Its interpretation depends on the differing label semantics, class supports, and acquisition conditions of the two collections.

\Needspace{24\baselineskip}
Table~\ref{tab:perclass} gives class-level metrics computed from the Dataset-A confusion matrix. Supported-class macro-F1 is 0.8493, weighted F1 is 0.8063, and balanced accuracy is 0.8302.

\begin{table}[H]\centering\small
\caption{Dataset-A metrics computed from the confusion matrix. The averaging set comprises the four supported regional classes; the final rows summarize predictions into other labels and the total error count.}
\label{tab:perclass}
\begin{tabular}{lrrrr}
\toprule
True label & Support & Precision & Recall & F1 \\
\midrule
Lamongan & 10 & 0.8889 & 0.8000 & 0.8421 \\
Malang & 16 & 0.9167 & 0.6875 & 0.7857 \\
Trenggalek & 6 & 0.7143 & 0.8333 & 0.7692 \\
Tulungagung & 1 & 1.0000 & 1.0000 & 1.0000 \\
\midrule
Supported-class mean & 33 & 0.8800 & 0.8302 & 0.8493 \\
\multicolumn{5}{l}{Predictions to B labels: Betawi 1, Dayak 1, Parang 1, Sekarjagad 1.} \\
\multicolumn{5}{l}{Other errors: four confusions among A labels. Total errors: 8/33.} \\
\bottomrule
\end{tabular}
\end{table}

The four additional predicted labels, Betawi, Dayak, Parang, and Sekarjagad, have zero ground-truth support in this subset. Including their four zero F1 values in the averaging set produces an eight-label macro-F1 of 0.4246, or 0.42 at two-decimal precision. These values illustrate the effect of the averaging inventory when a source subset receives predictions from a larger class space.

Tulungagung contributes a single test image, making its perfect score sensitive to individual outcomes. Appendix~\ref{app:labels} gives the Dataset-B per-class results at the two-decimal precision of the classification report.

\Needspace{18\baselineskip}
\subsection{Captioner adaptation and class-label retention}
Table~\ref{tab:caption} compares reference-based caption metrics across the six conditions. PFT-NCI achieves the highest scores on all six measures.

\begin{table}[H]\centering\scriptsize
\caption{Reference-based caption metrics on 150 Dataset-A images. Columns compare captioner adaptation and inference-time class injection under the evaluation protocol in Section~\ref{sec:evaluation}.}
\label{tab:caption}
\setlength{\tabcolsep}{3.5pt}
\begin{tabular}{lrrrrrr}
\toprule
&\multicolumn{2}{c}{Baseline}&\multicolumn{2}{c}{PFT}&\multicolumn{2}{c}{CIFT}\\
Metric & NCI & CI & NCI & CI & NCI & CI \\
\midrule
BLEU-1 & .4531 & .4438 & .6058 & .4490 & .4444 & .4447 \\
BLEU-4 & .2004 & .2639 & .4415 & .2675 & .2020 & .2698 \\
ROUGE-L & .4296 & .4674 & .6775 & .4696 & .4218 & .4761 \\
METEOR & .5419 & .5502 & .7307 & .5781 & .5424 & .5738 \\
SPICE & .3152 & .3701 & .5249 & .3761 & .3160 & .3714 \\
BERTScore & .9108 & .9147 & .9363 & .9102 & .9051 & .9110 \\
\bottomrule
\end{tabular}
\end{table}

\Needspace{17\baselineskip}
Table~\ref{tab:diagnostics} reports class-label mention and image--text compatibility. All three CI conditions attain MMR of 0.9067. Class-label mention is therefore invariant across the captioner variants under the injected prompt. This pattern is consistent with an interface that strongly encourages reproduction of the predicted label.

\begin{table}[H]\centering\scriptsize
\caption{Class-label mention and image--text compatibility on the caption-evaluation set. MMR measures ground-truth class-string inclusion; CLIPScore is shown in the evaluation's numerical scale.}
\label{tab:diagnostics}
\setlength{\tabcolsep}{3.5pt}
\begin{tabular}{lrrrrrr}
\toprule
&\multicolumn{2}{c}{Baseline}&\multicolumn{2}{c}{PFT}&\multicolumn{2}{c}{CIFT}\\
Diagnostic & NCI & CI & NCI & CI & NCI & CI \\
\midrule
MMR & .0000 & .9067 & .3800 & .9067 & .0000 & .9067 \\
CLIPScore & .3396 & .3375 & .3233 & .3303 & .3393 & .3310 \\
\bottomrule
\end{tabular}
\end{table}

CIFT-CI exceeds Baseline-CI in BLEU-4 by 0.0059, while its CLIPScore is lower by 0.0065. These small differences describe changes in reference agreement and image--text compatibility. Paired uncertainty estimates and independent attribute-level judgments would clarify their relationship to factual correctness.

PFT-NCI's leading reference-based scores in Table~\ref{tab:caption} accompany the lowest CLIPScore in Table~\ref{tab:diagnostics}. This disagreement shows that model rankings depend on the evaluation criterion. The qualitative examples provide a complementary view of the visual assertions and cultural identities expressed in the generated text.

\Needspace{43\baselineskip}
\subsection{Qualitative analysis of caption content and error propagation}
\begin{figure}[H]\centering\small
\setlength{\tabcolsep}{6pt}
\renewcommand{\arraystretch}{1.15}
\begin{tabularx}{\textwidth}{@{}p{2.3cm}>{\raggedright\arraybackslash}X@{}}
\toprule
Image and class & Caption excerpts by condition \\
\midrule
\includegraphics[width=2.0cm,height=2.6cm,keepaspectratio]{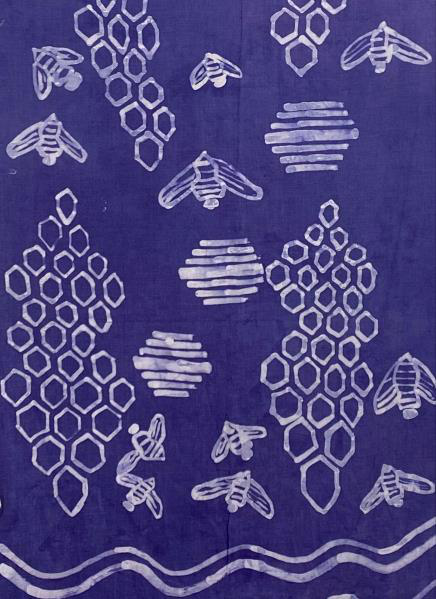}\newline Tasik Madu\newline\textbf{Malang}
& \textbf{PFT-NCI:} ``the malang motif''; ``lotus, bunga melati and sekartaji mask''\newline
\textbf{CIFT-NCI:} ``honeycomb motifs''; ``hexagons representing beehives and bees''\newline
\textbf{CIFT-CI:} ``the malang motif''; ``stylized bees and honeycomb structures'' \\
\addlinespace[9pt]
\includegraphics[width=2.0cm,height=2.6cm,keepaspectratio]{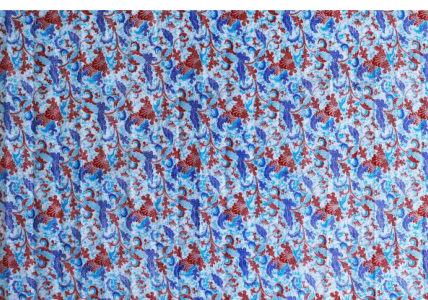}\newline Bunga Nirwana 2\newline\textbf{Tulungagung}
& \textbf{PFT-NCI:} ``the malang motif''; ``hibiscus''\newline
\textbf{CIFT-NCI:} ``swirling motifs''; ``leaves or vines''\newline
\textbf{CIFT-CI:} ``the trenggalek motif''; ``stylized leaves and flowers'' \\
\addlinespace[9pt]
\includegraphics[width=2.0cm,height=2.6cm,keepaspectratio]{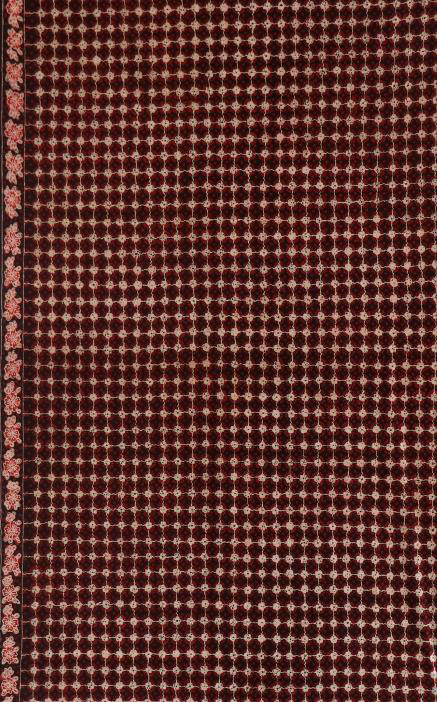}\newline Melati Deret\newline\textbf{Lamongan}
& \textbf{PFT-NCI:} ``the truntum motif''; ``stylized flowers or leaves''\newline
\textbf{CIFT-NCI:} ``small dot motifs''; ``tiny circles''\newline
\textbf{CIFT-CI:} ``the trenggalek motif''; ``traditional indonesian patterns'' \\
\bottomrule
\end{tabularx}
\caption{Generated caption excerpts across three model conditions. Tasik Madu illustrates correct identity retention under CI and unsupported visual nouns under PFT-NCI. Bunga Nirwana 2 and Melati Deret illustrate propagation of an incorrect Trenggalek prediction. The cases are selected to illustrate these mechanisms.}
\label{fig:qual}
\end{figure}

Figure~\ref{fig:qual} presents three image--caption cases across PFT-NCI, CIFT-NCI, and CIFT-CI. The excerpts permit direct comparison of class identity and visible attributes.

Tasik Madu visibly contains bees and honeycomb-like shapes, whereas the PFT-NCI output names lotus, jasmine, and a mask. Its CIFT-CI output includes the correct Malang identity together with bee and honeycomb descriptions. In the other two cases, CI reproduces the classifier's Trenggalek prediction despite different reference identities. The corresponding NCI excerpts use broader descriptions of flowers, leaves, dots, or layout.

These cases illustrate how class conditioning can preserve an identity or propagate an upstream error. Their selection supports analysis of the mechanisms involved, while limiting conclusions about the frequency of these behaviors across the evaluation set.

\subsection{Training dynamics with pseudo-caption supervision}
Training loss decreases from approximately 2.10 to 0.35 for adaptation using Dataset A alone and from 0.92 to 0.12 for adaptation using A+B. The curves show convergence on both training corpora. Differences in corpus composition and caption distributions limit interpretation of their absolute loss values as evidence of improved held-out caption quality.

\Needspace{25\baselineskip}
\subsection{Resource measurements}
Table~\ref{tab:resources} summarizes memory allocation, numerical status, and component latency. Model allocation is 5.4946 GB for NF4 and 9.6551 GB for the BF16/hybrid configuration, a difference of approximately 43\%. Peak allocations are 7.2415 GB and 13.1161 GB, respectively.

\begin{table}[H]\centering\small
\caption{Memory allocation and component latency. NF4 completes inference, while the BF16/hybrid configuration encounters a NaN failure. Latency is summarized as mean and standard deviation across 30 images.}
\label{tab:resources}
\begin{tabular}{lrr}
\toprule
Memory (GB) & NF4 & BF16 / hybrid \\
\midrule
Model allocation & 5.4946 & 9.6551 \\
Peak allocation & 7.2415 & 13.1161 \\
Numerical status & Completed & NaN failure \\
\midrule
\multicolumn{3}{l}{NF4 latency, mean $\pm$ SD seconds per image:} \\
Image I/O & \multicolumn{2}{c}{$0.2303\pm0.1026$} \\
Classifier & \multicolumn{2}{c}{$0.6225\pm0.3110$} \\
Captioner & \multicolumn{2}{c}{$21.7541\pm3.5514$} \\
End-to-end & \multicolumn{2}{c}{$22.6069\pm3.6892$} \\
\bottomrule
\end{tabular}
\end{table}

The BF16/hybrid run encounters a NaN failure and includes 134 tensors on a meta/offloaded path. Caption-quality comparison between the two loading configurations therefore requires a numerically stable BF16 run.

The NF4 cascade takes a mean of 22.6069 seconds per image, of which the captioner accounts for 21.7541 seconds, or approximately 96.23\%. Generation is thus the dominant component of the measured runtime. Decoding uses a ceiling of 256 new tokens, and image processing uses a 262,144-pixel cap.

\section{Discussion and Limitations}\label{sec:discussion}
All three CI variants attain MMR of 0.9067 despite their different adaptation conditions. This invariance is consistent with strong retention of the predicted class under the injected prompt. The qualitative cases demonstrate the dependency on classifier accuracy: the correct Malang prediction enters the Tasik Madu caption, while incorrect Trenggalek predictions enter the descriptions of Bunga Nirwana 2 and Melati Deret.

Label retention and visual factuality describe different aspects of caption quality. MMR measures the presence of a class string, while colors, ornaments, and spatial arrangement concern the visual assertions made about the fabric. Tasik Madu illustrates this distinction: the PFT-NCI caption includes the correct regional identity alongside lotus, jasmine, and mask descriptions that conflict with its visible bee and honeycomb forms. The three selected cases support analysis of such errors, while their selection limits inference about the prevalence of factual inaccuracies across the full evaluation set.

The disagreement between reference-based metrics and CLIPScore further qualifies the model rankings. PFT-NCI leads the reference-based measures while attaining the lowest image--text compatibility score. Because the reference captions share a recurring template, agreement in syntax and frequently repeated nouns can contribute to high lexical scores. Image--text compatibility provides a complementary view, although its interpretation depends on the embedding model's representation of local cultural terminology. These measures therefore capture distinct properties of the generated descriptions.

The independently constructed classifier and captioner partitions restrict conclusions about complete-cascade generalization. Some caption-evaluation images may have entered classifier training, and the overlap is unquantified. The caption results consequently characterize generation under the implemented partition scheme, with independence from both adapted components unresolved. The classifier's source-specific accuracy gap also combines differences in regional and motif-based taxonomies, class support, and acquisition conditions. Dataset A's 33-image classifier test, including a single Tulungagung image, further limits the stability of class-level estimates.

The A-only and A+B loss trajectories characterize adaptation to different caption corpora. Lower loss on the expanded corpus is compatible with easier prediction of its textual regularities and provides limited evidence about visual grounding. The contribution of pseudo-caption supervision to held-out generation remains unresolved by the loss comparison. This limits attribution of the six-condition caption results to the addition of Dataset B.

NF4 supports inference in the evaluated environment, while the BF16/hybrid NaN failure limits caption-quality comparisons. The 30-image latency sample characterizes runtime on the L40 and depends on generated length and image processing. Incomplete Dataset-B provenance and metric-configuration details constrain reproducibility. In collection-management workflows, generated captions serve as draft metadata for expert review of cultural attribution and visible attributes.

\section{Conclusion}
This study examines class-conditioned batik captioning through a DINOv2--Qwen2.5-VL cascade and a six-condition evaluation. Predicted-class injection yields a class-label mention rate of 0.9067 across all three captioner conditions. Source-specific classifier results show a 22.57-percentage-point accuracy gap, whose interpretation depends on the collections' different label inventories and training support. Class-level analysis further demonstrates the influence of the averaging convention on macro-F1.

Caption metrics produce different model rankings, and qualitative examples show both correct identity retention and propagation of classification errors. Together, these findings identify the classifier--captioner interface as a key point for evaluating cultural attribution. The results support assessing class-label retention alongside the factual grounding of the visual descriptions, within the generalization limits imposed by the component-specific data partitions.

\bibliographystyle{splncs04}
\bibliography{refs}
\clearpage\appendix\small
\section{Class Inventory and Per-Class Performance}\label{app:labels}
Dataset A comprises Lamongan, Malang, Trenggalek, and Tulungagung. Class indices are assigned by ordering the class labels; reproducing the mapping requires the original label encoding. Table~\ref{tab:bclasses} lists the 20 Dataset-B classes and their per-class results.

\begin{table}[H]\centering\small
\caption{Dataset-B precision, recall, and F1 at the two-decimal precision of the classification report. Each class contains 15 test images.}
\label{tab:bclasses}
\begin{tabular}{lrrr@{\quad}lrrr}
\toprule
Class & P & R & F1 & Class & P & R & F1 \\
\midrule
Betawi & 1 & 1 & 1 & Sidoluhur & 1 & .87 & .93 \\
Bokor Kencono & 1 & 1 & 1 & Sidomukti & 1 & 1 & 1 \\
Buketan & 1 & 1 & 1 & Sidomulyo & .88 & .93 & .90 \\
Dayak & .94 & 1 & .97 & Singa Barong & 1 & 1 & 1 \\
Jlamprang & .94 & 1 & .97 & Srikaton & 1 & 1 & 1 \\
Kawung & 1 & .93 & .97 & Tribusono & 1 & 1 & 1 \\
Liong & 1 & 1 & 1 & Tujuh Rupa & 1 & 1 & 1 \\
Mega Mendung & 1 & 1 & 1 & Truntum & 1 & 1 & 1 \\
Parang & 1 & 1 & 1 & Wahyu Tumurun & 1 & 1 & 1 \\
Sekarjagad & 1 & .93 & .97 & Wirasat & .94 & 1 & .97 \\
\bottomrule
\end{tabular}
\end{table}

\section{Pseudo-Caption Generation Prompt}\label{app:prompt}
The following instruction conditions pseudo-caption generation on each image's class label. The exact evaluation prompts for every NCI and CI condition remain unspecified.

\begin{lstlisting}
You are an expert batik annotator. Your task is to generate a precise, objective, and highly accurate visual description of the provided batik image.
This batik is classified as the '{kelas}' motif.
Instructions:
1. Start the first sentence with exactly: 'The batik fabric features the {kelas} motif, characterized by [list 2-4 dominant motif colors] motifs.'
   List only the 2-4 most visually dominant motif colors. Do NOT enumerate all visible colors.
2. In the second sentence, identify the dominant motif elements (e.g. fern plants, fish, geometric shapes, floral patterns).
3. In the third sentence, describe how the motifs are arranged across the fabric (e.g. non-geometric, diagonal, repeating grid, symmetrical). Include spacing, density, and orientation if clearly visible.
4. If isen-isen (small dots, short lines, or fillers) are clearly visible, include them.
5. Describe only what is directly visible in the image. Do NOT invent colors not clearly present. Do NOT include cultural meanings, historical context, or symbolic interpretations.
6. Output strictly ONE coherent paragraph of 3-4 sentences, without any introductory or concluding remarks.
\end{lstlisting}

\end{document}